\documentclass[conference]{IEEEtran}
\IEEEoverridecommandlockouts
\usepackage{cite}
\usepackage{amsmath,amssymb,amsfonts}
\usepackage{algorithm}
\usepackage{algpseudocode}
\usepackage{graphicx}
\usepackage{textcomp}
\usepackage{xcolor}
\usepackage{subfigure}
\newtheorem{example}{Example}

\def\BibTeX{{\rm B\kern-.05em{\sc i\kern-.025em b}\kern-.08em
    T\kern-.1667em\lower.7ex\hbox{E}\kern-.125emX}}
    
\begin{document}
\title{Conceptual In-Context Learning and Chain of Concepts: Solving Complex Conceptual Problems Using Large Language Models
}

\author{\IEEEauthorblockN{ Nishtha N. Vaidya\textsuperscript{1,2}, Thomas A. Runkler\textsuperscript{1,2}, Thomas Hubauer\textsuperscript{1}, Veronika Haderlein-Hoegberg\textsuperscript{1}, Maja Milicic Brandt\textsuperscript{1}}
\IEEEauthorblockA{\textit{\textsuperscript{1} Siemens AG, Munich, Germany \textsuperscript{2} Technical University of Munich, Germany} \\
Corresponding author - \textit{nishtha.vaidya@siemens.com}}}
% \author{\IEEEauthorblockN{Anonymous Authors}}

%reasoning is the process of solving a problem and conceptual problem is a a spicific type of problem
\IEEEoverridecommandlockouts
\IEEEpubid{\makebox[\columnwidth]{978-1-5386-5541-2/18/\$31.00~\copyright2024 IEEE \hfill}
\hspace{\columnsep}\makebox[\columnwidth]{ }}

\maketitle
\IEEEpubidadjcol
\begin{abstract}
    Science and engineering problems fall in the category of complex conceptual problems that require specific conceptual information (CI) like math/logic-related know-how, process information, or engineering guidelines to solve them. Large Language Models (LLMs) are promising agents to solve such complex conceptual problems due to their implications in advancing engineering and science tasks like assisted problem solving. But vanilla LLMs, trained on open-world data, lack the necessary CI. In this work, we specifically explore shallow customization methods (SCMs) of LLMs for solving complex conceptual problems. We propose two novel SCM algorithms for LLMs, to augment LLMs with CI and enable LLMs to solve complex conceptual problems: Conceptual In-Context Learning (C-ICL) and Chain of Concepts (CoC). The problem tackled in this paper is the generation of proprietary data models in the engineering/industry domain based on conceptual information in data modelling guidelines. We evaluate our algorithms on varied sizes of the OpenAI LLMs against four evaluation metrics related to syntactic and semantic correctness, time and cost incurred. The proposed algorithms perform better than currently popular LLM SCMs like In-context Learning (ICL) and Chain of Thoughts (CoT). It was observed that as compared to CoT, response correctness increased by 30.6\%  and 29.88\%  for the new SCMs C-ICL and CoC, respectively. Qualitative analysis suggests that the proposed new SCMs activate emergent capabilities in LLMs, previously unobserved in the existing SCMs. They make the problem solving processes more transparent and reduce hallucinations and the tendency of model responses to copy examples from prompts (parroting).
\end{abstract}


\section{Introduction}
   Large Language Models (LLMs) have revolutionized solving everyday problems like text summarising, copy-writing and knowledge acquisition~\cite{openai2024gpt4technicalreport,lin2023usinglanguagemodelsknowledge}. LLMs can solve these general problems through the model's base knowledge, captured by its large representation power trained on amounts of open-world data. Many problems that LLMs can solve are never introduced to the LLM during training. These \textit{emergent capabilities} of LLMs are enabled through specialized natural language (NL) prompting techniques like In-Context Learning (ICL)~\cite{lu2024emergentabilitieslargelanguage, hahn2023theoryemergentincontextlearning, chen2023understanding} and Chain of Thoughts (CoT)~\cite{wei2023chainofthoughtpromptingelicitsreasoning} referred to as shallow customization methods (SCM) of LLMs as they do not alter model embeddings. Therefore LLMs, with or without such SCM, are good at general problem-solving or formal linguistic abilities. Formal linguistic competency tasks are based on syntactic rules and statistical patterns that characterize a language. On the flip side, open-world data often lack specialized information like math/logic related know-how, process information or engineering guidelines and hence, LLMs lack such knowledge too. It has been shown that LLMs often fail to solve even simple math problems~\cite{nezhurina2024alicewonderlandsimpletasks} and have inconsistencies in their knowledge when reasoning~\cite{ucedasosa2024reasoningconceptsllmsinconsistencies}. 
   
    Problem solving is a complex process that humans excel through the interplay of several aspects of their cognitive abilities~\cite{Pizlo2022-PIZPSC,WUSTENBERG20121}. It is an acquired skill for humans, often requiring reasoning~\cite{sternberg1980reasoning}. It is both logical and creative in nature, hence challenging for LLMs. While LLMs do well on formal linguistic competency tasks, LLM's base knowledge lacks essential problem solving mechanisms that humans possess~\cite{WANG201081}, and cannot solve functional linguistic competency problems~\cite{mahowald2024dissociatinglanguagethoughtlarge} well. LLMs struggle at even problems with single correct answers~\cite{cheng2024inductivedeductiverethinkingfundamental}.%—the ability to use language in real-world situations
    
    Cognitive science theories prove \textit{concepts are key to problem solving}~\cite{gardenfors2023reasoning}. The term concept used is as in cognitive science literature~\cite{Laurence1999-LAUCAC-3} meaning ‘building blocks of thought'. Various problem solving mechanisms like deductive, abductive, and inductive reasoning rely on conceptual information (CI). We call such problems that can be solved with CI as \textit{complex conceptual problems (CP)}.  Many science and engineering problems fall in this category of complex CP with often multiple correct answers. LLMs as agents to solve such CPs~\cite{shen2024tagllmrepurposinggeneralpurposellms, collins2022structuredflexiblerobustbenchmarking} are crucial due to its implications in advancing engineering and science problems~\cite{hardy2023large}. It has been proven that LLMs solve such \textit{complex CP} through spurious correlations from training data and often generate false positives (hallucinate)~\cite{williams2024easyproblemsllmswrong,
     Tang_2023, nezhurina2024alicewonderlandsimpletasks}. Hence (plain) LLMs are not well suited for solving complex CPs~\cite{frieder2023mathematicalcapabilitieschatgpt, kalyanpur2024llmarcenhancingllmsautomated, davoodi2024llmsintelligentthinkersintroducing}. 
     
     In this paper we specifically explore SCMs of LLMs for solving complex CPs. For this problem, the deep customisation of LLMs (like fine-tuning~\cite{zhang2024scalingmeetsllmfinetuning, ha2024fusiondomainadaptedvisionlanguage} and LoRA~\cite{DBLP:journals/corr/abs-2106-09685}) is not feasible due to data sparsity and over-fitting risks.
    %\todo{is reasoning problem, functional linguistic problems, CP clear} 
    We show that currently popular SCM for LLMs are unsuccessful in solving complex CPs. The models were observed to hallucinate or parrot (copy examples from SCM prompts), indicating missing CI. In cases where SCM-based LLMs succeeds, LLMs are not robust as they generates correct responses through incorrect problem solving mechanisms. In the second part, we provide two novel SCMs for augmenting LLMs with CI. We show the success of our methods on a simple logical reasoning CP with a single correct answer and a complex engineering CP with multiple correct answers. The main paper contributions are 1. We demonstrate the ineffectiveness of current SCMs for complex CP solving and provide a hypothesis of why they are ineffective, 2. We propose two new SCM algorithms to augment LLMs with CI (C-ICL and CoC) leading to successful complex CP solving, 3. We apply the proposed SCMs to a real-world industrial problem of generating proprietary data models (PDM), 4. We provide an evaluation study of the SCMs and provide a method to evaluate multiple aspects of response correctness to highlight problem solving skills gained by LLMs and their receptiveness to SCM.
    %~\footnote{The pseudocode for the proposed methods is shown in Algorithm~\ref{alg:CoC}. The actual code cannot currently be made public due to product integration plans.}
\section{Existing Methodology}
    %\todo{merge II, III, IV in single section}
    % \subsection{Alice In Wonderland Problem} 
    For intuitive understanding of the different SCMs and to demonstrate the need for more sophisticated approaches to solve CPs using LLMs we use the following example 
        \begin{example}
            \textit{If Alice has X sisters and Y brothers.
            How many sisters does Alice's brother have?}
        \end{example}
        % \begin{figure}[h]
        %     \centering
        %     \includegraphics[scale=0.4]{fig/method-Page-3.drawio.pdf}
        %     \caption{Alice in Wonderland Problem : Calculating number of sisters of Alice's brother if Alice has 3 sisters and 7 brothers.}
        %     \label{fig:aiw-problem}
        % \end{figure}
        The answer to this problem is $X$ + 1, as Alice herself needs to be counted when calculating the total number of sisters for a male sibling of Alice. But as \cite{nezhurina2024alicewonderlandsimpletasks} showcases LLMs perform poorly on this CP called as Alice in Wonderland (AIW).\\
        For e.g., for $X = 3$ and $Y = 7$, the correct response is $4$.
        \subsection{Simple Prompt}
\begin{figure*}[h]
        \centerline{\includegraphics[scale = 0.3]{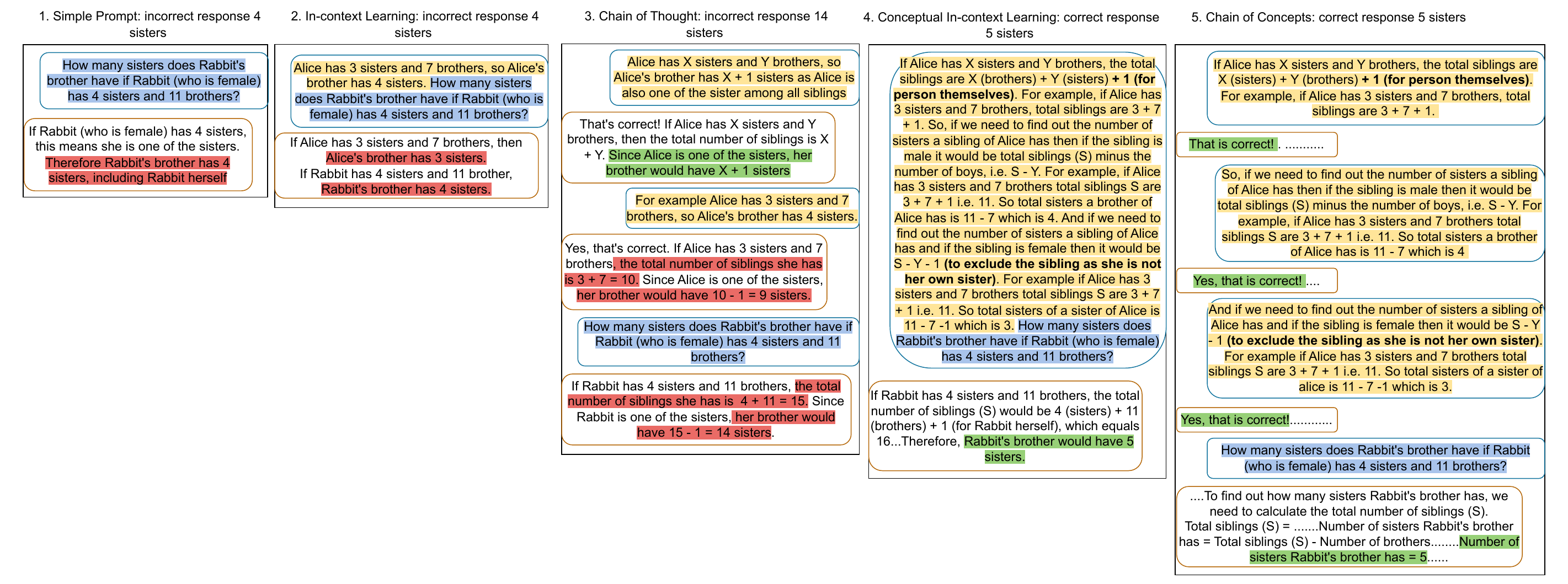}}
        \caption{The LLM response is customised through 5 different SCMs using GPT-3.5-turbo-16k. Figure markings - right-aligned box are model input, left-aligned are model response, SCM prompts are marked in yellow, CP in blue, and for model's response: correct are marked in green and incorrect in red. }
        \label{fig:AIW}
    \end{figure*}

         When normally prompted (no customisation to LLM) as in Fig.~\ref{fig:AIW}.1 ($X = 4$ and $Y = 11$), the response is incorrect: $4$.\\
         The model fails to add 1 and this clearly shows the poor capabilities of LLM as reasoners for such a simple problem. 
    \subsubsection{General Framework}
    \label{Sec:gen}
        Consider a CP $\mathcal{C}$ that can be solved with CI and an LLM $\mathcal{M}$ which takes as input an NL query and returns an NL response. In general interaction, a user with a CP $q$ pertaining to $\mathcal{C}$ and obtains a model response $r$. The function approximator to model $\mathcal{M}$ is $m$, such that
        \begin{equation}
            r = m_{\mathcal{M}}(q)
        \end{equation}
        The interaction can be qualified by following metrics
        \paragraph{Response Correctness $\theta_{correct}$}
            We define response as correct when it matches the ground truth/ correct response
            \begin{equation}
                \theta_{C} = \sum_{i=1}^{N} \frac{k_i}{N}
            \end{equation}
            for $N$ response samples and the correctness of the $i^{th}$ sample
            \begin{equation}
              k_{i}= \begin{cases}
                1, & \text{if $r_i=r^*$}.\\
                0, & \text{otherwise}.
              \end{cases}
              \label{eqn:correct}
            \end{equation}
            where $r_i$ is $i^{th}$ model response and $r^*$ is the correct response. A complex CP can have different aspects of correctness like syntactic and semantic correctness $\{\theta_{C_{i}}\}$ (e.g., in Section~\ref{sec:exp}).
        \paragraph{Response Cost $\theta_{cost}$}
            Response cost is the cost of generating an LLM response to the CP. It is a parameter of the LLM and the token size (function of total characters in the query and response). Higher token size corresponds to higher response cost. Let $f_\mathcal{M}$ be the model's characteristic function to calculate the token size then
            \begin{equation}
                \theta_{cost} = f_{\mathcal{M}}(q+r)
                \label{eqn:token}
            \end{equation}
        \paragraph{Time Consumption $\theta_{time}$} If it takes time t to generate a response to given prompt then
            \begin{equation}
                \theta_{time} = t \text{ (wall-clock seconds)}
            \label{eqn:time}
            \end{equation} 
        \paragraph{Model Confidence $\theta_{P}$}
            If we utilise an SCM $\lambda$ (like Simple Prompt), then model confidence is
        \begin{equation}
            \theta_{P} = \alpha^{\lambda}_{\mathcal{M},r}
            \label{eqn:confidence}
        \end{equation}
        where $\alpha$ is the model's characteristic function (average of probabilities obtained from model for all response tokens) to determine the confidence for the generated response $r$.
    
    \subsection{In-context learning}
    \label{Sec:ICL}
        ICL is a one-shot prompting method. The query consists of the CP and an example $\epsilon$: $q+\epsilon$.
        For e.g., on giving an example of Alice with CP for Rabbit (Fig.~\ref{fig:AIW}.2) the response is incorrect: $4$.
    
    \subsection{Chain of Thoughts}
    \label{Sec:COT}
        %edit COT in the diagram, you should also give answer of LLM to be correct?} - no we supply the ansswer as ICL
        Introduced by \cite{wei2023chainofthoughtpromptingelicitsreasoning}, CoT is a few-shot prompting method in which the problem is broken down into a series of steps that lead to the correct answer. In some works, CoT is also implemented with a simple addition of "let's think step-by-step" to the end of the ICL prompt~\cite{NEURIPS2022_8bb0d291}. In other works, CoT is implemented by adding an explanation of reasoning and using examples, followed by CP.  
        
       We apply CoT consisting of an explanation of the solution through the example method to AIW as in Fig.~\ref{fig:AIW}.3, LLM again gives an incorrect answer.
        
        Formalising, let set  $\{(q_i, r_i)\}$ represent an $i^{th}$ query-response pair in the few shot CoT, then model input is
        \begin{equation}
            q_{n} = q \| (\|_{n=0}^{N-1}(q_i,r_i)), (\|\text{ : concatenation operation})
        \end{equation}
        for $N$ number of query-response pair and, response is
        
        \begin{equation}
            r = m_{\mathcal{M}}(q_{n})
        \end{equation}
        
         Then response cost and time taken become
         
         \begin{equation}
             \theta_{cost} = f_{\mathcal{M}}(q_n + r_n)
         \end{equation}
         
         \begin{equation}
             \theta_{time} = \sum_{i=1}^{N+1}(t_i)
         \end{equation}
        The $\theta_{P}$ are same as in %\todo{Do not use Eqn but (...)}
        (\ref{eqn:confidence}). 
        %\todo{model confidence also sum or?} - no we consider for only final response to the actual query
        
        The LLM cannot solve the AIW variation CP of number of sisters of a sister of Alice instead of brother. For example, following up the CoT conversation chain (Fig.~\ref{fig:AIW}.3) with
        \begin{example}
        \label{ex:1}
            \textit{'How many sisters does Rabbit's sister have?'}
        \end{example}
        The response is incorrect and non-numerical - "\textit{If Rabbit has X sisters and Y brothers, and one of those sisters is Rabbit's sister, then Rabbit's sister would have X - 1 sisters. This is because Rabbit herself is also counted as one of the sisters.}"
        
        %We have demonstrated how the currently known SCMs perform poorly on AIW.
    \section{Chain of Concepts And Conceptual ICL}
    The poor performance of LLMs on AIW demonstrates that the LLM lacks correct problem solving mechanisms. These mechanisms are the CI important to solve the CP. CI is missing in CoT or ICL where instead the steps to obtain the solution or examples are provided to the LLM. We hypothesise LLMs augmented with CI can solve CP and provide two new methods 
    \subsection{Conceptual In-context Learning}
        Consider we have a CI (we present a method to construct CI in Section~\ref{sec:CI}). A naive approach will be to construct a single prompt consisting of the CI and CP and provide it to LLM as a zero-shot prompt.
        Applying this to AIW as in Fig.~\ref{fig:AIW}.4 we get the correct answer. The response also consists of the reasoning mechanism explanation. 
        The metrics of this C-ICL method are calculated similar to that of plain ICL. %(Section~\ref{Sec:ICL}).
        \\
        %consistentency of correct answer higher\\
        \textit{Example Of Conceptual Information for AIW:}
    %\todo{reverse the order: first AIW then generalization - done now} - done now %\todo{remove notations}
        The input prompt in Fig.~\ref{fig:AIW}.4 shows the CI for AIW. This CI consists of introduction of calculation of the total number of siblings and explanation of addition of $1$ to $X + Y$, and an example. Next, the concept of role of the gender of a sibling is introduced by an explanation of calculating the sister of a male or female sibling, followed by examples and the CP.
    \subsection{Chain of Concepts}
         For CPs that require multiple concepts, instead of a single input prompt as in C-ICL, concepts can instead be incrementally introduced to the model in multiple inter-related prompts.  This method CoC when applied to LLMs for AIW (Fig.~\ref{fig:AIW}.5), leads to correct response and a more granular explanation of LLM's problem solving mechanisms.
        % It is also observed that LLM "self-corrects" based on the incrementally provided CI. These behaviors make the LLM robust and reliable when encountering CP variations. 
        We calculate the metrics for CoC in the same way as CoT.
        
        %\todo{add number of sisters of Alice's sister calculation, how concepts are not complete}
    Further on querying the LLM with AIW variation (Example~\ref{ex:1}) we get correct responses for both C-ICL and CoC. %\todo{add response}

    As compared to AIW which has a single correct answer, the complex CPs often have multiple correct responses (e.g., storytelling~\cite{see2019massivelypretrainedlanguagemodels}) making evaluation a challenge and requiring human reviewers. In Section~\ref{sec:exp} we apply SCMs and evaluated SCMs on a more complex CP. The algorithm to construct CoC and C-ICL is formalised as 
    \subsubsection{Extending General Framework}
        \paragraph{Conceptual Information As A Directed-Acyclic Graph}
             \begin{figure}[bp]
             \vspace{-1em}
             \centering
             \subfigure{1.}{\includegraphics[width=0.25\textwidth]{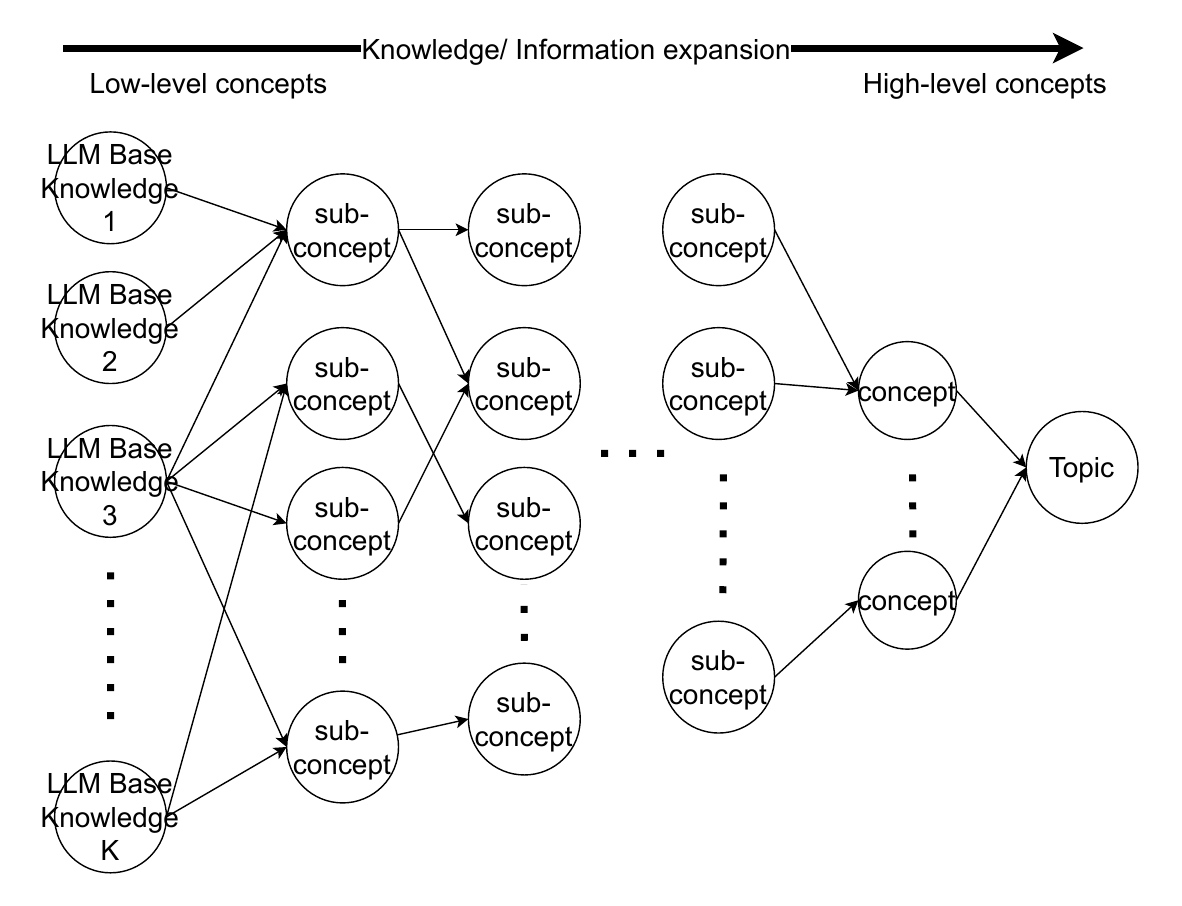}}
             \subfigure{2.}{\includegraphics[width=0.2\textwidth]{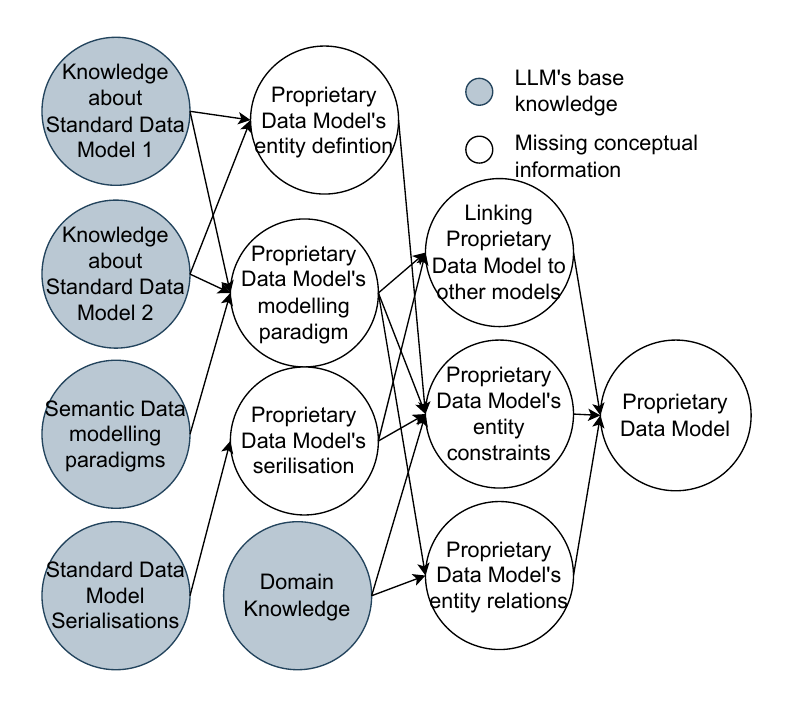}}
                \caption{Proposed SCM CoC based formulation of CI as a DAG: 1. General DAG of CI 2. CI DAG Example for proprietary data model generation CP.}
             \label{fig:topic}
            \end{figure}
            %\todo{add scale of higher and lower concepts to diagram}
            %\todo{do we introduce it as a DAG, unidirectional information flow, data source is flowing in unidirectional way, but as you traverse along the direction of these arrows the IE of the topic becomes more abstract}
            Let's think of CI $\mathcal{A}$ (the information required to solve the CP $\mathcal{C}$) as a unidirectional flow of information starting from the \textit{general knowledge}, called \textit{low-level concepts} that are more fundamental pieces of information, to the \textit{topic} which we define as a high-level concept that is a more abstract piece of information. The levels of abstraction represent different models of the same idea but with varying amounts of details~\cite{Floridi2004-FLOLAT}. We characterize $\mathcal{A}$ as a directed-acyclic graph (DAG) (Fig.~\ref{fig:topic}.1), where each \textit{node} $A$ represents a \textit{concept}. Each concept can be represented by \textit{lower}-level concepts or \textit{sub-concepts}. The \textit{edges} $E$ of the DAG link the sub-concepts to the concepts. The \textit{direction of the edge} shows the increasing abstractness of the \textit{concept}. Therefore, traversing along the DAG, the nodes that do not have any incoming edges are \textit{general knowledge} and the final node from which no edge originates is the \textit{topic} to which the CP belongs. A \textit{topic} is the most abstract information of the CI while \textit{general knowledge} is the least abstract. Therefore, $\mathcal{A}$ is a set of all nodes and edges $a \in A$ and $e \in E$
            \begin{equation}
                \mathcal{A} = \{A, E\}
            \end{equation}

        \paragraph{LLM's Knowledge}
        \label{sec:llmkg}
        Several research works~\cite{radford2018improving, DBLP:journals/corr/abs-2005-14165} have shed light on the fact that LLMs possess \textit{general knowledge} and \textit{low-level concepts} which we refer as base knowledge. They have shown that the LLMs can solve impressive arrays of problems with base knowledge~\cite{lu2024emergentabilitieslargelanguage}. These categories of problems are formal linguistic competence. While for the complementary problem of functional linguistic competence like reasoning, LLMs do not perform well~\cite{mahowald2024dissociatinglanguagethoughtlarge,nezhurina2024alicewonderlandsimpletasks} (e.g., AIW). These problems require LLMs to possess knowledge of \textit{complex CI} that is often not encountered in the open-world data. The base knowledge is specific to the LLM, its size and the training data used to it. The training data of an LLM might include some functional linguistic competency problems. However, many of the higher-level concepts essential for problem solving do not occur in the training data~\cite{mahowald2024dissociatinglanguagethoughtlarge}. Thus, LLMs lack the complete CI and cannot solve CPs, as seen in failure for AIW. However, LLM's knowledge of low-level concepts is an essential building-block to understand high-level concepts, making LLMs a powerful tool for solving CPs. 
        %\todo{replace by algorithm everywhere}
        \subsubsection{Algorithm To Construct Chain Of Concepts}
        \label{sec:CoC}
        \paragraph{Formulating The CI For CoC}
        \label{sec:CI}
            The DAG $\mathcal{A}$ has to be constructed by the domain expert/user for each CP. For a more complex CP the $\mathcal{A}$ required for any agent to solve the $\mathcal{C}$ gets bigger. This means the nodes, i.e. concepts, required can be more in number and more abstract, and/or the edges, i.e. relations among sub-concepts, are higher in number. A naive way to formulate $\mathcal{A}$ would be to provide all concepts and their relations to the model. However, this is not a feasible task as the low-level concepts are often large in number and models have limited token size. But as discussed in Section~\ref{sec:llmkg}, as LLMs possess base knowledge we omit low-level concepts and only reference them in relation to high-level concepts (edges of DAG). The knowledge of LLM is then a DAG $\mathcal{A}^* \subset \mathcal{A}$
            \begin{equation}
                \mathcal{A}^* = \{A^*, E^*\}
            \end{equation} 
            $\mathcal{A}^*$ of LLM is determined heuristically after $\mathcal{A}$ is constructed. The method to do this is to query the LLM for low-level concepts and identify the consistently correct responses. $\mathcal{A}_c$ is the difference between $ \mathcal{A}^*$ and $ \mathcal{A}$, and therefore consists of fewer nodes (concepts). Thus, we obtain LLM's missing CI $\mathcal{A}_c$. More optimised methods to formulate $\mathcal{A}_c$ can be adopted, but this is beyond the scope of current work.\\
            For an example of a CI for a proprietary data model generation (PDM) the CP (Example~\ref{ex:3}) is given in Fig.~\ref{fig:topic}.2: the LLM base knowledge (gray nodes) includes information like standard data models and high-level concepts (white nodes) include information like the PDM schema (unavailable openly).
            \begin{example}
                \textit{"Can you generate a $<$PDM name$>$ for $<$domain$>$?"} 
                \label{ex:3}
            \end{example}
        \paragraph{Constructing CoC}
            \begin{algorithm}
                \caption{Algorithm for Chain of Concepts}
                \label{alg:CoC}
                \begin{algorithmic}
                \Require $\mathcal{A}, \mathcal{M}, \mathcal{A}^*, P $\Comment{$P$ denotes generating NL prompt} 
                \State $ \mathcal{A}_c \gets \mathcal{A} \setminus \mathcal{A}^*$
                \State CoC = []
                \While{${A}_c \neq \emptyset$}
                \State $(a, e) \gets {\text{BFS}({A}_c )}$ \Comment{BFS: breadth first search, $a$: child node}
                \State $q \gets P((a,e))$ 
                \State $r \gets m_\mathcal{M}(q+\epsilon)$ 
                \State $\mathcal{A}^* \cup (a,e)$ %\Comment{$(a,e)$ node, edge}
                \State ${A}_c \gets \mathcal{A} \setminus \mathcal{A}^*$
                \State $\text{CoC} \gets \text{CoC}\text{.append}((q, r))$
                \EndWhile
                \State \Return CoC
                \end{algorithmic}
            \end{algorithm}
            The naive method to augment LLMs with the CI is converting the $\mathcal{A}_c$ to a single NL prompt and provide it to the LLM before the CP (\textit{C-ICL} method). Alternatively $\mathcal{A}_c$ is split into multiple prompts based of different concept nodes, which is CoC. The splitting is started from the low-level concept and relations, and sequentially relating them and introducing high-level concepts while providing the examples wherever possible (inspired from the success of few-shot prompting~\cite{li2024chain,madaan2022text}). This incremental way of prompting the concepts is equivalent to traversing the DAG $\mathcal{A}_c$ in breadth-first search manner. The algorithm is explained in Algorithm~\ref{alg:CoC} and an example of the DAG of the CP constructed using it is shown in Fig.~\ref{fig:topic}.2 for the CP in Example~\ref{ex:3}. This many-shot prompting method CoC builds CI incrementally, reducing the ambiguity in LLM's problem solving mechanisms.
\section{Application To Data Model Engineering}
    Data Models are machine-readable structured data representations that relate entities and their relationships according to syntactic rules based on schemata and semantic rules given by the domain. The process of operating on various types of data models is called Data Modeling (DM). DM has widespread applications in industries like defining assets and creating digital twins. In open-world data, many standard formalisms for DM (e.g., AAS~\cite{bader2019semantic}, OWL~\cite{antoniou2009web}) exist. 
        %\todo{In section V.A.1 I think we should add a high level explanation what our problem in data modeling actually is, so that the average NLP expert is able to understand.}
        % \subsubsection{LLMs As A Solution For  Generation}
    Many companies build PDMs tailored to a specific industry and application, and these are unavailable openly. The domain and DM experts are tasked with the challenge of PDM generation. These challenges are
    \begin{enumerate}
        \item Current SoTA of PDM generation is manual drafting by DM experts. It incurs high cognitive burden requiring knowledge acquisition of the domain, expertise of DM and PDM modelling paradigms, and learning DM tools.
        \item It is time consuming and requires multiple iterations.
        \item Unavailability of tools for new PDMs makes accessibility and adoption of PDM a challenge.
    \end{enumerate}
    LLMs with the base knowledge of the standard data models and DM, formal linguistic competence and NL user interface have potential to solve PDM CPs. However, as LLMs do not possess CI of PDMs, LLMs without any customisation cannot do PDM generation. We comparatively evaluate different SCMs for this task in the next section.
\section{Experiments}
    \label{sec:exp}
    \subsection{Experimental Setup}
\begin{figure*}[h]
        \centerline{\includegraphics[scale = 0.3]{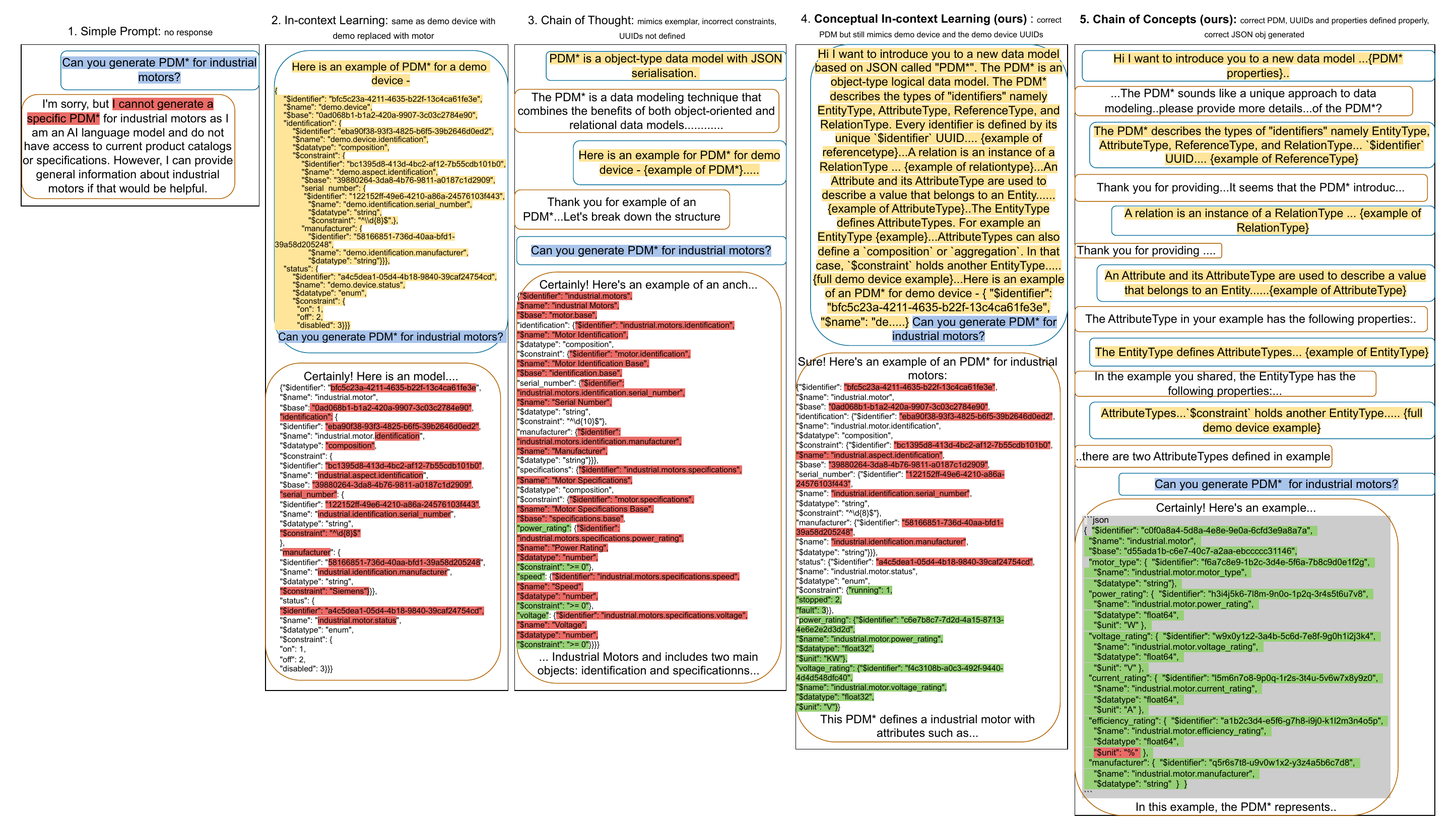}}
        \caption{Sample response using GPT-3.5-turbo-16k for \textit{PDM*} generation using five SCMs. The left aligned boxes are LLM's response and right ones are input prompts. Colour markings mean: SCM - yellow, CP - blue, errors in the generated PDM - red and correct features - green, JSON generated - grey. }
        \label{fig:PDM}
    \end{figure*}

   The experimental parameters are detailed in Table~\ref{tab1}. The customised LLMs are queried with the CP in Example~\ref{ex:3} for $\beta$ randomly chosen domains. The target \textit{PDM*} (based on a company data model) has a JSON serialisation. $\gamma$ samples of $\beta$ domains for each of the $3$ models of varied sizes and for $5$ different SCMs are generated to create a dataset of $495$ \textit{PDM*s}. Fig.~\ref{fig:topic}.2 shows the CI constructed using CoC algorithm. In Fig.~\ref{fig:PDM} a sample that is representative of the dataset for industrial motor domain for the SCMs is shown. 
   %The full data cannot currently be made public due to company intellectual property reasons. 
   The PDM CP is chosen to demonstrate the complexity of real-world CPs that are unavailable openly, require manual evaluation, and have multiple correct answers. As we evaluate the data manually the dataset size is chosen to optimise effort and produce meaningful results. The evaluation metrics are as defined previously and $\theta_{correct}$ comprises of 4 aspects 
   \begin{itemize}
       \item  Valid JSON check $\theta_{C_1}$: the model generates a \textit{PDM*} that can be parsed to a valid JSON file.
       \item Syntactic check $\theta_{C_2}$: the valid JSON is passing all syntactic rules of the \textit{PDM*} schema.
       \item Parroting check $\theta_{C_3}$: indicates lack of "creative" or "original" generations which is a essential for PDM.
       \item Quasi-semantic check $\theta_{C_4}$: evaluation becomes challenging for this complex CP with multiple correct responses, hence in the current scope we do a quasi-semantic check, defined as average of the 8 KPIs (correctly defined entity 1. names, 2. base, 3. identifiers, 4. constraints, 5. composition, 6. units, 7. datatypes, and 8. relationships, defined with help of an DM expert. 
   \end{itemize}
   The quasi-semantic check is the most crucial correctness aspect for the PDM CP as if it is not semantically meaningful then it is invalid, irrespective of its syntactic correctness. 
    \begin{table}[bp]\vspace{-2em}
        \caption{Experimental Parameters}
        \begin{center}
        \begin{tabular}{p{0.2\linewidth}p{0.6\linewidth}}
        \hline 
        \textbf{Parameters}&\textbf{Values}\\
        \hline \hline
        SCM & Simple Prompt, ICL, CoT, C-ICL and CoC \\ 
        \hline
        Models & GPT-3.5-turbo-16k, GPT-4-16k, GPT-4-turbo-128K-1106\\ \hline
        Hyperparameters & default except maximum overall tokens $= 16000$, maximum response tokens $ = 4096$\\ \hline
        No. of CP ($\beta$), Samples ($\gamma$) &  $\beta = 3$ (domains - coffee making, industrial motors, real estate buildings), $\gamma = 11$\\
        \hline 
        %\multicolumn{4}{l}{$^{\mathrm{a}}$Sample of a Table footnote.}
        \end{tabular}
        \label{tab1}
        \end{center}
    \end{table}
    \begin{figure*}
    \centering
    \subfigure{\includegraphics[width=0.22\textwidth]{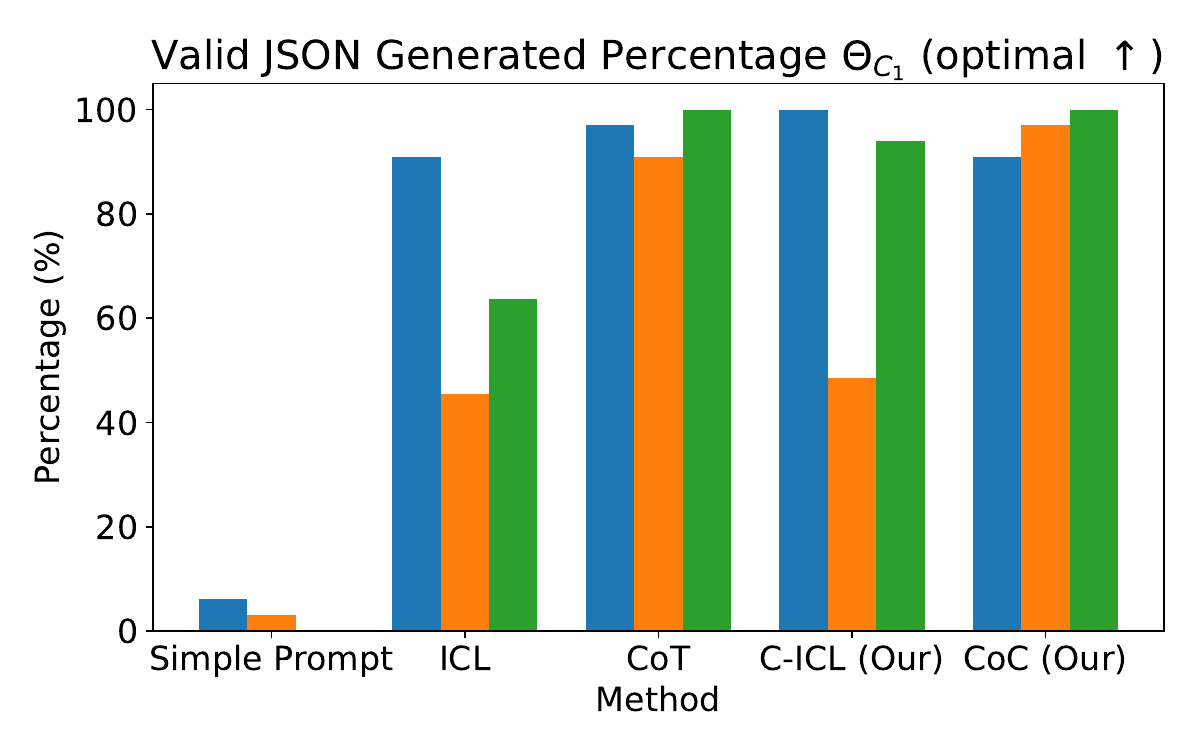}}
    \subfigure{\includegraphics[width=0.22\textwidth]{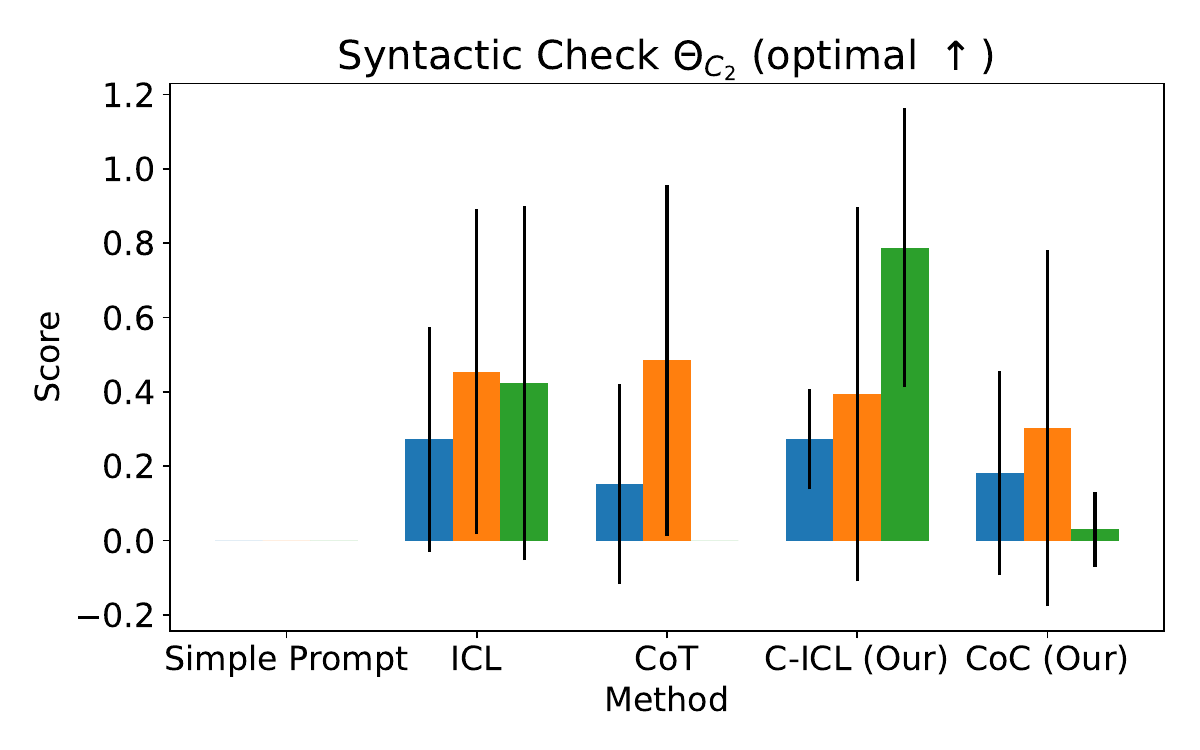}}
    \subfigure{\includegraphics[width=0.22\textwidth]{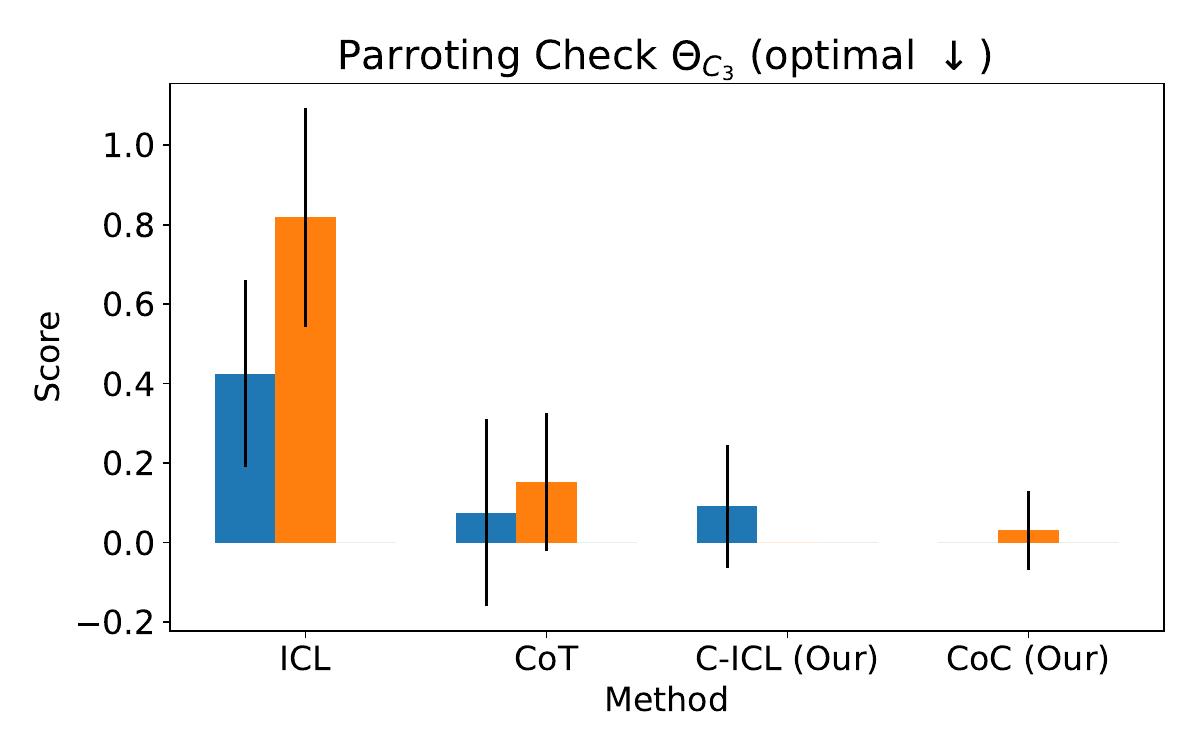}}
    \subfigure{\includegraphics[width=0.22\textwidth]{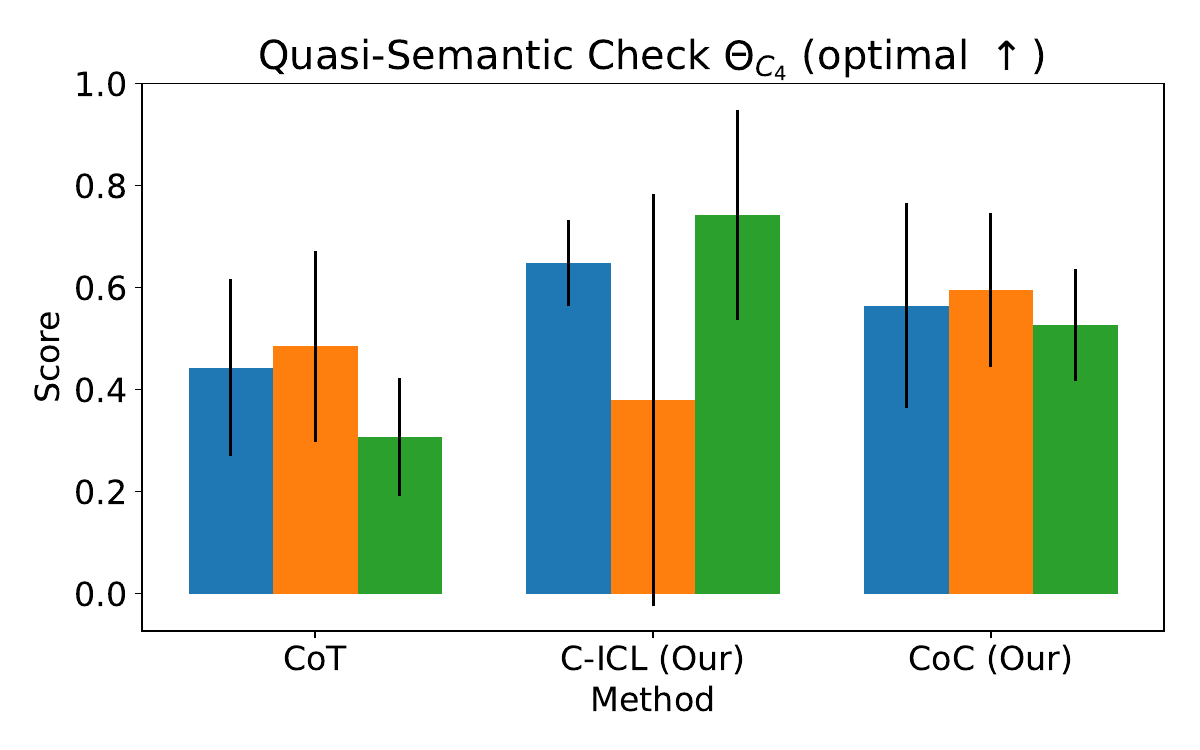}}
    \subfigure{\includegraphics[width=0.22\textwidth]{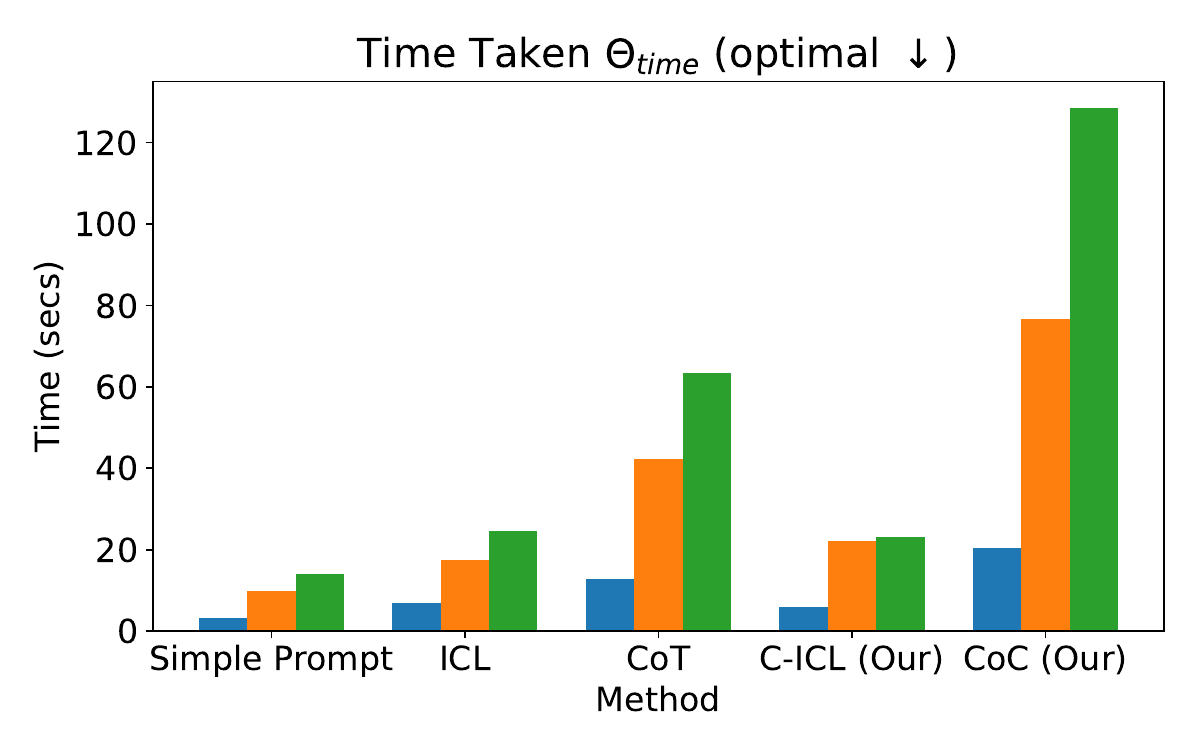}} 
    \subfigure{\includegraphics[width=0.22\textwidth]{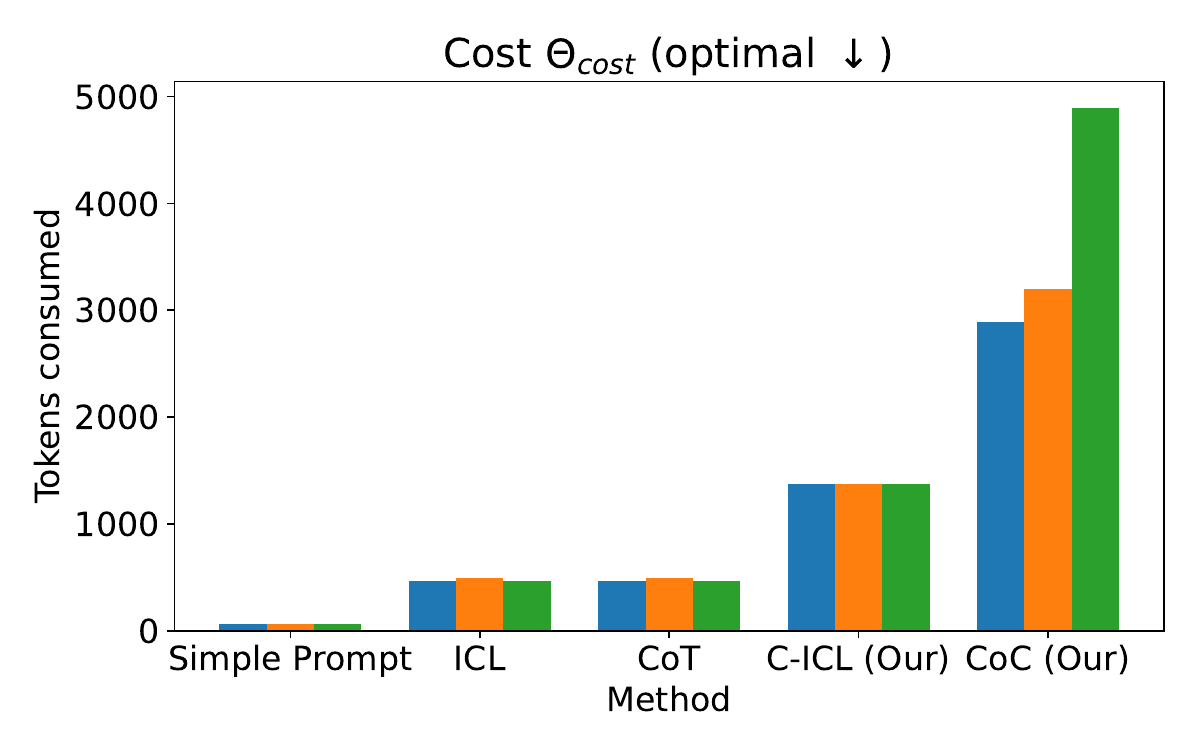}} 
    \subfigure{\includegraphics[width=0.33\textwidth]{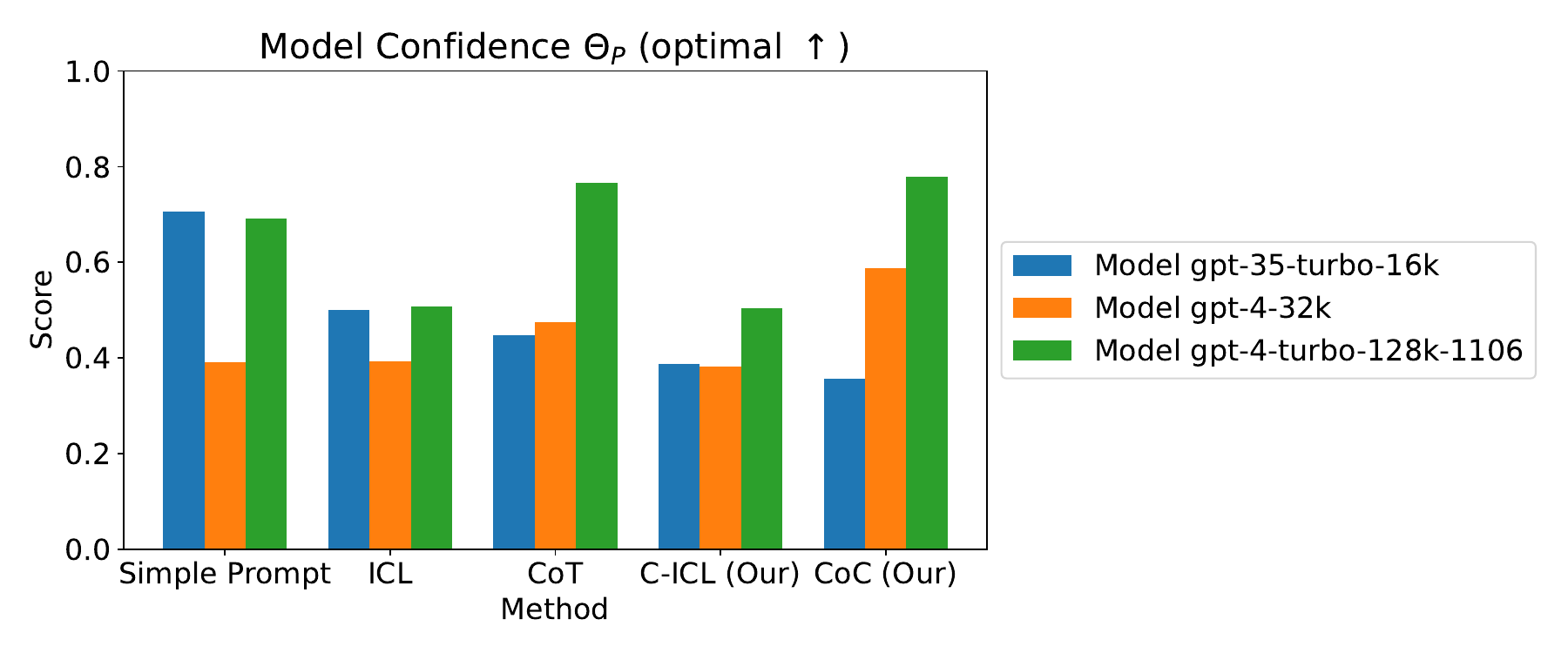}}  
    \caption{ For the experiment matrix (Table~\ref{tab1}) the metrics are plotted (from left to right and top to bottom)  - $\theta_{C_1}$, $\theta_{C_2}$, $\theta_{C_3}$, $\theta_{C_4}$, $\theta_{time}$, $\theta_{cost}$, $\theta_{P}$}.
    \label{fig:metrics}
    \end{figure*}
\subsection{Experimental Results, Discussion And Related Works}
    \subsubsection{Quantitative evaluation}
    In Fig.~\ref{fig:metrics} each bar represents a metric averaged over $\gamma \times \beta $ samples for a model and a SCM. The trends are - \textit{$\theta_{C_1}$:} CoC and CoT performed best for all model sizes, C-ICL showed steep decrease in performance for mid-size model, ICL performed comparatively very poor, and Simple Prompt performed the worst with $\approx 0$ valid JSONs. \textit{$\theta_{C_2}$:} C-ICL and ICL performed the best with comparable results, followed by CoC and CoT with comparable results and both perform significantly poorly for larger models, Simple Prompt performed worst. \textit{$\theta_{C_3}$:} ICL suffered from parroting significantly, rendering it invalid for this particular CP, minor parroting was observed for lower size model with CoT/C-ICL/CoC, C-ICL and CoC are highly robust to parroting with none observed for larger models, Simple Prompt was not evaluated as it did not generate any valid \textit{PDM*}. \textit{$\theta_{C_4}$:} CoC performs best with low variance and consistently across all models, comparatively C-ICL suffers heavy variance and also inconsistent across model sizes, CoT performs consistently but poorer than C-ICL and CoC, ICL is not evaluated due to parroting.
    \textit{$\theta_{time}$ and $\theta_{cost}$:} CoC and C-ICL took highest time and incurred highest costs, increasing proportionately with model size. \textit{$\theta_{P}$:} methods that performed poor in correctness showed high confidence indicating hallucination, like for Simple Prompt $\theta_{C1}$ is close to $0$ but $\theta_{P}$ is $0.7$. 
    \subsubsection{Qualitative evaluation} 1. Simple Prompt - responses are either "cannot do the task" or hallucinated imaginary PDM (Fig.~\ref{fig:PDM}.1).
    2. ICL - high parroting, most responses are same as example \textit{PDM*} with only attribute names changed as per the query domain (Fig.~\ref{fig:PDM}.2).
    3. CoT - higher parroting (Fig.~\ref{fig:PDM}.3) as compared to CoC/C-ICL, suffers from hallucinations like adding new entities to \textit{PDM*} schema.
    4. C-ICL - more conservative model confidence vs. correctness which can be interpreted as low hallucination, low in robustness compared to CoC as variance is high in $\theta_C$, some responses contain comprehensive attribute definition (Fig.~\ref{fig:PDM}.4) unobserved in ICL/CoT.
    5. CoC - more consistent explanation of generated PDM in response, defines relationships, and better in comprehensive attributes definition (Fig.~\ref{fig:PDM}.5) as compared to C-ICL.
    \subsubsection{Discussion} As compared to CoT the combined quasi-semantic correctness with non-parroting increased by 30.6\% and 29.88\% for C-ICL and CoC respectively. The generated PDMs were of higher semantic and syntactic quality and suffer extremely low hallucination and parroting.
    \textit{CoC vs. C-ICL:} CoC gives the best quality. In time/cost constrained scenarios C-ICL with human validation can be a viable alternative. The choice also depends on CI size and correctness aspect. For simple CPs with few concepts like AIW, C-ICL works well. In cases where the CP has multiple correct answers and concepts like PDM, incrementally introducing concepts through CoC has shown advantages, similar to pedagogical teaching of concepts to humans. \textit{Model choice:} Large model is essential when CI is large, or cost is high. For CoC, smaller model is also a good choice as it shows comparable performance to the best performing larger model and reduces energy consumption. The large non-turbo GPT models performed poorly in CP solving as compared to any sized turbo (improved training). \textit{CI:} Keeping the CI concise showed better performance with lower cost. \textit{Prompt engineering:} As a learning from the experiments when converting the CI to NL prompts, the best responses are generated when the prompts are explicit and pedagogical. \textit{Transparency benefits:} The proposed SCMs generated responses with elaboration of problem solving mechanisms, which can enable auto-validation of responses.
    \subsubsection{Related Works} 
Works like~\cite{bao2024llmschainofthoughtnoncausalreasoners} show that CoT is poor at solving CPs. Similar to~\cite{NEURIPS2023_271db992,Besta_2024} our methods improve over CoT where instead of left-to-right token based decision making our methods enable LLM to base it on a collection of related prompts. The difference is between their exploratory curation of collection of thoughts which are more suitable for serial decision-making problems like solving puzzles vs. our structured CI and fixed curation of \textit{concepts} to solve complex CPs effectively. Most works for LLMs for problem solving are limited to math reasoning CPs~\cite{davoodi2024llmsintelligentthinkersintroducing, li2024chain} as opposed to closed-domain complex CPs that are practically more relevant. %~\cite{shen2024tagllmrepurposinggeneralpurposellms} employ deep customisation for complex CP solving but as discussed apriori they face scalability challenge. 
Domain knowledge injection with CoT is explored in~\cite{jiang2024llmsmathematicalreasoningmistakes} as an alternative to RAG~\cite{lewis2021retrievalaugmentedgenerationknowledgeintensivenlp} but have not shown to augment LLMs with problem solving skills nor target complex CPs. Similarly, ~\cite{wang2024retaskrevisitingllmtasks} provides a theoretical model using CoT applied to sentence prediction task. But they do not focus on complex CP or problem-solving skills, nor do extensive evaluation (like for hallucinations). The distinguishing factor of our work is that we provide a systematic algorithm enabling LLM to solve complex CP and a comprehensive evaluation strategy.

\section{Conclusion and Future Direction}    
    To the best of the knowledge of the authors this is the first paper providing model-agnostic SCM of LLMs for solving complex CPs. We demonstrated that our proposed SCMs (C-ICL and CoC) solved a real-world complex data engineering CP successfully. The proposed SCMs show a significant improvement over existing SCMs reducing hallucination and parroting issues. They bring forth emergent problem solving capabilities that have never been explicitly included in CI.
    %and also demonstrate capability to tackle modifications to CP. For e.g., on following up CoC customised LLM for PDM generation problem with \textit{"Can you convert $<$PDM$>$ to OWL?"} led to correct DM transformation.
    The introduced SCM CoC performed the best, with potential to generate meaningful first-cut solutions for complex engineering and science tasks. Directions for future work include improving syntactic correctness of CoC, applying CoC to other complex CPs and to multimodal LLMs with multimodal CI.

%%%%%%%%%%%%%%%%%%%%%%%%%%%%%%%%%%%%%%%%%%%%%%%%%%%%%%%%%%%%
\bibliographystyle{unsrt}
\bibliography{cite}

\begin{thebibliography}{10}

\bibitem{openai2024gpt4technicalreport}
OpenAI, Josh Achiam, Steven Adler, Sandhini Agarwal, Lama Ahmad, Ilge Akkaya,
  Florencia~Leoni Aleman, Diogo Almeida, Janko Altenschmidt, and et~al.
  Sam~Altman.
\newblock {GPT-4 Technical Report}, 2024.

\bibitem{lin2023usinglanguagemodelsknowledge}
Fangzhen Lin, Ziyi Shou, and Chengcai Chen.
\newblock {Using Language Models For Knowledge Acquisition in Natural Language
  Reasoning Problems}, 2023.

\bibitem{lu2024emergentabilitieslargelanguage}
Sheng Lu, Irina Bigoulaeva, Rachneet Sachdeva, Harish~Tayyar Madabushi, and
  Iryna Gurevych.
\newblock {Are Emergent Abilities in Large Language Models just In-Context
  Learning?}, 2024.

\bibitem{hahn2023theoryemergentincontextlearning}
Michael Hahn and Navin Goyal.
\newblock {A Theory of Emergent In-Context Learning as Implicit Structure
  Induction}, 2023.

\bibitem{chen2023understanding}
Shuo Chen, Zhen Han, Bailan He, Mark Buckley, Philip Torr, Volker Tresp, and
  Jindong Gu.
\newblock {Understanding and Improving In-Context Learning on Vision-language
  Models}.
\newblock {\em arXiv preprint arXiv:2311.18021}, 1(2), 2023.

\bibitem{wei2023chainofthoughtpromptingelicitsreasoning}
Jason Wei, Xuezhi Wang, Dale Schuurmans, Maarten Bosma, Brian Ichter, Fei Xia,
  Ed~Chi, Quoc Le, and Denny Zhou.
\newblock {Chain-of-Thought Prompting Elicits Reasoning in Large Language
  Models}, 2023.

\bibitem{nezhurina2024alicewonderlandsimpletasks}
Marianna Nezhurina, Lucia Cipolina-Kun, Mehdi Cherti, and Jenia Jitsev.
\newblock {Alice in Wonderland: Simple Tasks Showing Complete Reasoning
  Breakdown in State-Of-the-Art Large Language Models}, 2024.

\bibitem{ucedasosa2024reasoningconceptsllmsinconsistencies}
Rosario Uceda-Sosa, Karthikeyan~Natesan Ramamurthy, Maria Chang, and Moninder
  Singh.
\newblock {Reasoning about concepts with LLMs: Inconsistencies abound}, 2024.

\bibitem{Pizlo2022-PIZPSC}
Zygmunt Pizlo.
\newblock {\em {Problem Solving: Cognitive Mechanisms and Formal Models}}.
\newblock Cambridge University Press, 2022.

\bibitem{WUSTENBERG20121}
Sascha Wüstenberg, Samuel Greiff, and Joachim Funke.
\newblock {Complex problem solving — More than reasoning?}
\newblock {\em Intelligence}, 40(1):1--14, 2012.

\bibitem{sternberg1980reasoning}
Robert~J Sternberg.
\newblock {\em {Reasoning, problem solving, and intelligence}}.
\newblock Canada Institute for Scientific and Technical Information, 1980.

\bibitem{WANG201081}
Yingxu Wang and Vincent Chiew.
\newblock {On the cognitive process of human problem solving}.
\newblock {\em Cognitive Systems Research}, 11(1):81--92, 2010.
\newblock Brain Informatics.

\bibitem{mahowald2024dissociatinglanguagethoughtlarge}
Kyle Mahowald, Anna~A. Ivanova, Idan~A. Blank, Nancy Kanwisher, Joshua~B.
  Tenenbaum, and Evelina Fedorenko.
\newblock {Dissociating language and thought in large language models}, 2024.

\bibitem{cheng2024inductivedeductiverethinkingfundamental}
Kewei Cheng, Jingfeng Yang, Haoming Jiang, Zhengyang Wang, Binxuan Huang,
  Ruirui Li, Shiyang Li, Zheng Li, Yifan Gao, Xian Li, Bing Yin, and Yizhou
  Sun.
\newblock {Inductive or Deductive? Rethinking the Fundamental Reasoning
  Abilities of LLMs}, 2024.

\bibitem{gardenfors2023reasoning}
Peter G{\"a}rdenfors and Mat{\'\i}as Osta-V{\'e}lez.
\newblock {Reasoning with concepts: A unifying framework}.
\newblock {\em Minds and Machines}, 33(3):451--485, 2023.

\bibitem{Laurence1999-LAUCAC-3}
Stephen Laurence and Eric Margolis.
\newblock Concepts and cognitive science.
\newblock In Eric Margolis and Stephen Laurence, editors, {\em {Concepts: Core
  Readings}}, pages 3--81. MIT Press, 1999.

\bibitem{shen2024tagllmrepurposinggeneralpurposellms}
Junhong Shen, Neil Tenenholtz, James~Brian Hall, David Alvarez-Melis, and
  Nicolo Fusi.
\newblock {Tag-LLM: Repurposing General-Purpose LLMs for Specialized Domains},
  2024.

\bibitem{collins2022structuredflexiblerobustbenchmarking}
Katherine~M. Collins, Catherine Wong, Jiahai Feng, Megan Wei, and Joshua~B.
  Tenenbaum.
\newblock {Structured, flexible, and robust: Benchmarking and improving large
  language models towards more human-like behavior in out-of-distribution
  reasoning tasks}, 2022.

\bibitem{hardy2023large}
Mathew Hardy, Ilia Sucholutsky, Bill Thompson, and Tom Griffiths.
\newblock {Large language models meet cognitive science: LLMs as tools, models,
  and participants}.
\newblock In {\em Proceedings of the annual meeting of the cognitive science
  society}, volume~45, 2023.

\bibitem{williams2024easyproblemsllmswrong}
Sean Williams and James Huckle.
\newblock {Easy Problems That LLMs Get Wrong}, 2024.

\bibitem{Tang_2023}
Ruixiang Tang, Dehan Kong, Longtao Huang, and Hui Xue.
\newblock {Large Language Models Can be Lazy Learners: Analyze Shortcuts in
  In-Context Learning}.
\newblock In {\em Findings of the Association for Computational Linguistics:
  ACL 2023}. Association for Computational Linguistics, 2023.

\bibitem{frieder2023mathematicalcapabilitieschatgpt}
Simon Frieder, Luca Pinchetti, Alexis Chevalier, Ryan-Rhys Griffiths, Tommaso
  Salvatori, Thomas Lukasiewicz, Philipp~Christian Petersen, and Julius Berner.
\newblock {Mathematical Capabilities of ChatGPT}, 2023.

\bibitem{kalyanpur2024llmarcenhancingllmsautomated}
Aditya Kalyanpur, Kailash~Karthik Saravanakumar, Victor Barres, Jennifer
  Chu-Carroll, David Melville, and David Ferrucci.
\newblock {LLM-ARC: Enhancing LLMs with an Automated Reasoning Critic}, 2024.

\bibitem{davoodi2024llmsintelligentthinkersintroducing}
Arash~Gholami Davoodi, Seyed Pouyan~Mousavi Davoudi, and Pouya Pezeshkpour.
\newblock {LLMs Are Not Intelligent Thinkers: Introducing Mathematical Topic
  Tree Benchmark for Comprehensive Evaluation of LLMs}, 2024.

\bibitem{zhang2024scalingmeetsllmfinetuning}
Biao Zhang, Zhongtao Liu, Colin Cherry, and Orhan Firat.
\newblock {When Scaling Meets LLM Finetuning: The Effect of Data, Model and
  Finetuning Method}, 2024.

\bibitem{ha2024fusiondomainadaptedvisionlanguage}
Cuong~Nhat Ha, Shima Asaadi, Sanjeev~Kumar Karn, Oladimeji Farri, Tobias
  Heimann, and Thomas Runkler.
\newblock {Fusion of Domain-Adapted Vision and Language Models for Medical
  Visual Question Answering}, 2024.

\bibitem{DBLP:journals/corr/abs-2106-09685}
Edward~J. Hu, Yelong Shen, Phillip Wallis, Zeyuan Allen{-}Zhu, Yuanzhi Li,
  Shean Wang, and Weizhu Chen.
\newblock {LoRA: Low-Rank Adaptation of Large Language Models}.
\newblock {\em CoRR}, abs/2106.09685, 2021.

\bibitem{NEURIPS2022_8bb0d291}
Takeshi Kojima, Shixiang~(Shane) Gu, Machel Reid, Yutaka Matsuo, and Yusuke
  Iwasawa.
\newblock {Large Language Models are Zero-Shot Reasoners}.
\newblock In S.~Koyejo, S.~Mohamed, A.~Agarwal, D.~Belgrave, K.~Cho, and A.~Oh,
  editors, {\em Advances in Neural Information Processing Systems}, volume~35,
  pages 22199--22213. Curran Associates, Inc., 2022.

\bibitem{see2019massivelypretrainedlanguagemodels}
Abigail See, Aneesh Pappu, Rohun Saxena, Akhila Yerukola, and Christopher~D.
  Manning.
\newblock {Do Massively Pretrained Language Models Make Better Storytellers?},
  2019.

\bibitem{Floridi2004-FLOLAT}
Luciano Floridi and J.~W. Sanders.
\newblock {Levellism and the Method of Abstraction}.
\newblock In {\em IEG Research Report}. 2004.

\bibitem{radford2018improving}
Alec Radford.
\newblock {Improving language understanding by generative pre-training}.
\newblock 2018.

\bibitem{DBLP:journals/corr/abs-2005-14165}
Tom~B. Brown, Benjamin Mann, Nick Ryder, and et. al.
\newblock {Language Models are Few-Shot Learners}.
\newblock {\em CoRR}, abs/2005.14165, 2020.

\bibitem{li2024chain}
Zhiyuan Li, Hong Liu, Denny Zhou, and Tengyu Ma.
\newblock {Chain of thought empowers transformers to solve inherently serial
  problems}.
\newblock {\em arXiv preprint arXiv:2402.12875}, 2024.

\bibitem{madaan2022text}
Aman Madaan and Amir Yazdanbakhsh.
\newblock {Text and patterns: For effective chain of thought, it takes two to
  tango}.
\newblock {\em arXiv preprint arXiv:2209.07686}, 2022.

\bibitem{bader2019semantic}
Sebastian~R Bader and Maria Maleshkova.
\newblock {The Semantic Asset \\Administration Shell}.
\newblock In {\em Semantic Systems. The Power of AI and Knowledge Graphs: 15th
  International Conference, SEMANTiCS 2019, Karlsruhe, Germany, September
  9--12, 2019, Proceedings 15}, pages 159--174. Springer, 2019.

\bibitem{antoniou2009web}
Grigoris Antoniou and Frank~van Harmelen.
\newblock Web ontology language: Owl.
\newblock {\em Handbook on ontologies}, pages 91--110, 2009.

\bibitem{bao2024llmschainofthoughtnoncausalreasoners}
Guangsheng Bao, Hongbo Zhang, Linyi Yang, Cunxiang Wang, and Yue Zhang.
\newblock {LLMs with Chain-of-Thought Are Non-Causal Reasoners}, 2024.

\bibitem{NEURIPS2023_271db992}
Shunyu Yao, Dian Yu, Jeffrey Zhao, Izhak Shafran, Tom Griffiths, Yuan Cao, and
  Karthik Narasimhan.
\newblock {Tree of Thoughts: Deliberate Problem Solving with Large Language
  Models}.
\newblock In A.~Oh, T.~Naumann, A.~Globerson, K.~Saenko, M.~Hardt, and
  S.~Levine, editors, {\em Advances in Neural Information Processing Systems},
  volume~36, pages 11809--11822. Curran Associates, Inc., 2023.

\bibitem{Besta_2024}
Maciej Besta, Nils Blach, Ales Kubicek, Robert Gerstenberger, Michal
  Podstawski, Lukas Gianinazzi, Joanna Gajda, Tomasz Lehmann, Hubert
  Niewiadomski, Piotr Nyczyk, and Torsten Hoefler.
\newblock {Graph of Thoughts: Solving Elaborate Problems with Large Language
  Models}.
\newblock {\em Proceedings of the AAAI Conference on Artificial Intelligence},
  38(16):17682–17690, March 2024.

\bibitem{jiang2024llmsmathematicalreasoningmistakes}
Zhuoxuan Jiang, Haoyuan Peng, Shanshan Feng, Fan Li, and Dongsheng Li.
\newblock {LLMs can Find Mathematical Reasoning Mistakes by Pedagogical
  Chain-of-Thought}, 2024.

\bibitem{lewis2021retrievalaugmentedgenerationknowledgeintensivenlp}
Patrick Lewis, Ethan Perez, Aleksandra Piktus, Fabio Petroni, Vladimir
  Karpukhin, Naman Goyal, Heinrich Küttler, Mike Lewis, Wen tau Yih, Tim
  Rocktäschel, Sebastian Riedel, and Douwe Kiela.
\newblock {Retrieval-Augmented Generation for Knowledge-Intensive NLP Tasks},
  2021.

\bibitem{wang2024retaskrevisitingllmtasks}
Zhihu Wang, Shiwan Zhao, Yu~Wang, Heyuan Huang, Jiaxin Shi, Sitao Xie, Zhixing
  Wang, Yubo Zhang, Hongyan Li, and Junchi Yan.
\newblock {Re-TASK: Revisiting LLM Tasks from Capability, Skill, and Knowledge
  Perspectives}, 2024.

\end{thebibliography}
\end{document}